\documentclass[11pt]{article}

\usepackage{acl}

\usepackage{enumitem}
\usepackage{times}
\usepackage{latexsym}
\usepackage[T1]{fontenc}
\usepackage[utf8]{inputenc}
\usepackage{microtype}
\usepackage{inconsolata}

\usepackage{graphicx}
\usepackage{booktabs}
\usepackage{amsmath}
\usepackage{amssymb}
\usepackage{url}
\usepackage{hyperref}

\usepackage{listings}
\title{Broken Chains: The Cost of Incomplete Reasoning in LLMs}

\author{
Ian Su \quad Gaurav Purushothaman \quad Jey Narayan \quad Ruhika Goel \quad Kevin Zhu \quad \\
\textbf{Sunishchal Dev \quad Yash More \quad Maheep Chaudhary}
}

\begin{document}
\maketitle



\begin{abstract}
Reasoning-specialized models like OpenAI's 5.1 and DeepSeek-V3.2 allocate substantial inference compute to extended chain-of-thought (CoT) traces, yet reasoning tokens incur significant costs. How do different reasoning modalities of code, natural language, hybrid, or none do perform under token constraints?
We introduce a framework that constrains models to reason exclusively through code, comments, both, or neither, then systematically ablates token budgets to 10\%, 30\%, 50\%, and 70\% of optimal. We evaluate four frontier models (GPT-5.1, Gemini 3 Flash, DeepSeek-V3.2, Grok 4.1) across mathematical benchmarks (AIME, GSM8K, HMMT).
Our findings reveal: (1) \textbf{truncated reasoning can hurt} as DeepSeek-V3.2 achieves 53\% with no reasoning but only 17\% with truncated CoT at 50\% budget; (2) \textbf{code degrades gracefully} as Gemini's comments collapse to 0\% while code maintains 43-47\%; (3) \textbf{hybrid reasoning underperforms} single modalities; (4) \textbf{robustness is model-dependent} as Grok maintains 80-90\% at 30\% budget where OpenAI and DeepSeek collapse to 7-27\%. These results suggest incomplete reasoning chains actively mislead models, with implications for deploying reasoning-specialized systems under resource constraints.
\end{abstract}


\section{Introduction}

Chain-of-thought (CoT) reasoning has become the dominant paradigm for complex problem-solving with large language models~\citep{wei2022chain}. By generating intermediate reasoning steps, models achieve substantial improvements on mathematical and logical tasks~\citep{kojima2022large}. This success has driven the development of reasoning-specialized models such as OpenAI's 5.1~\citep{openai2025gpt51}, DeepSeek-V3.2~\citep{deepseekai2025deepseekv32}, and others, that allocate inference compute to extended reasoning traces, sometimes generating thousands of tokens before producing an answer.

\begin{figure}
    \centering
    \includegraphics[width=\linewidth]{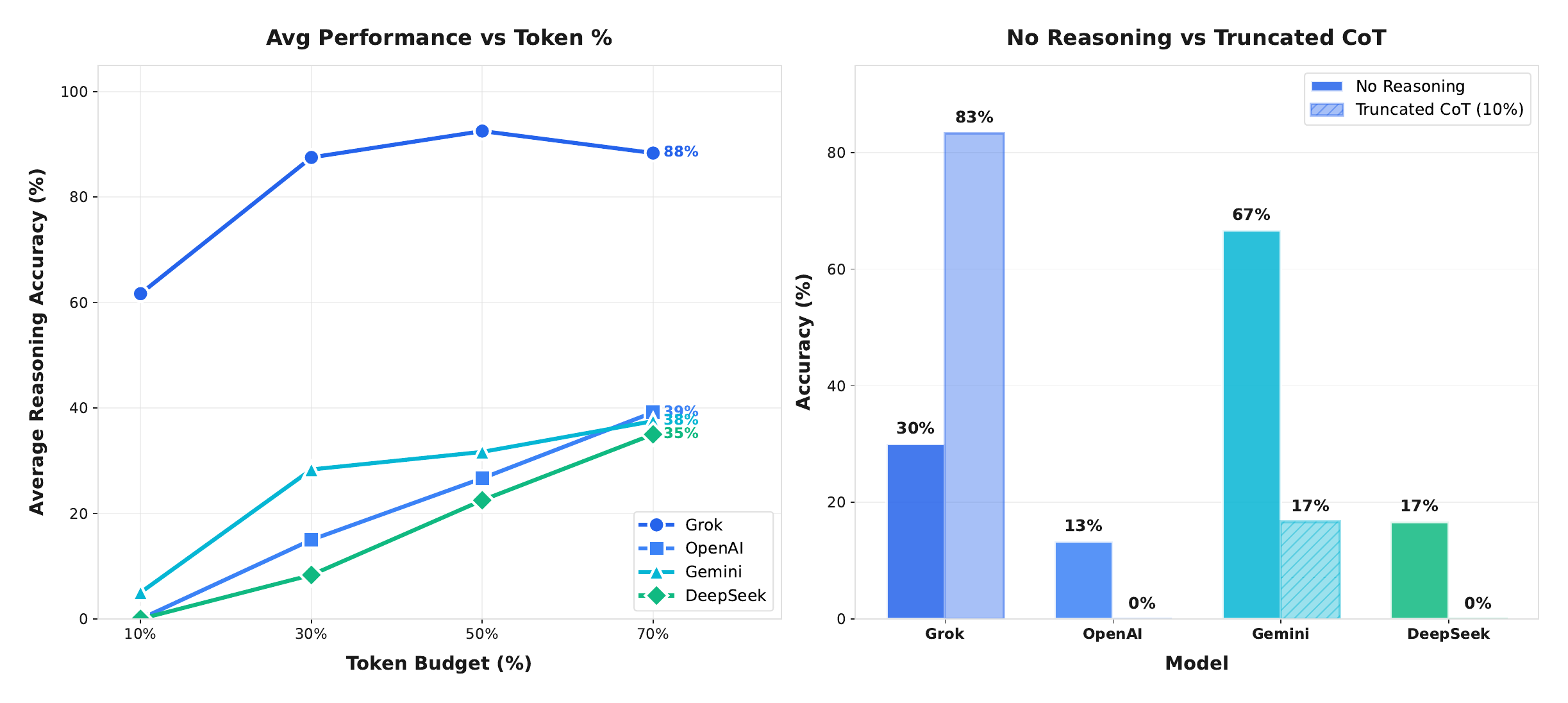}
    \caption{Reasoning robustness under token constraints. (a) Average accuracy across reasoning conditions as token budget varies from 10\% to 70\% of optimal. Grok maintains high performance even at severe constraints, while other models degrade substantially. (b) Comparing no explicit reasoning versus truncated CoT at 10\% budget reveals a striking paradox: for Gemini, GPT-5.1, and DeepSeek, \textit{no reasoning outperforms truncated reasoning}, suggesting incomplete chains actively mislead models. Only Grok benefits from truncated CoT.}
    \label{fig:main}
\end{figure}

However, this paradigm rests on an implicit assumption: \textit{more explicit reasoning is always beneficial}. As Figure~\ref{fig:main} illustrates, this assumption breaks down under token constraints as truncated reasoning can actually \textit{hurt} performance, with Gemini achieving 67\% accuracy with no reasoning but only 17\% with truncated CoT. In practice, reasoning tokens constitute significant inference costs, and real-world deployments often impose token constraints due to latency, cost, or context limitations. This raises a critical question: \textbf{how do different reasoning modalities of code, natural language, or hybrid do perform under token budgets, and what happens when reasoning chains are truncated?}

\paragraph{Challenges.} Investigating reasoning modality robustness presents methodological difficulties. Standard reasoning traces \textit{entangle multiple formats} for interleaving natural language, symbolic manipulation, and code, making it difficult to isolate contributions. The relationship between token budget and performance is \textit{non-linear and model-dependent}. Furthermore, \textit{no established methodology} exists for constraining reasoning to specific modalities while preserving task validity.

\paragraph{Our Approach.} We introduce a controlled framework that constrains models to reason through five conditions: (1) \textit{code-only} is reasoning as executable code; (2) \textit{comments-only} is natural language without code; (3) \textit{both} is hybrid; (4) \textit{nothing} is direct answer without explicit reasoning; and (5) \textit{standard CoT}. We systematically ablate token budgets to 10\%, 30\%, 50\%, and 70\% of optimal across four frontier models, including the reasoning-specialized DeepSeek-V3.2 on mathematical benchmarks (AIME, GSM8K, HMMT).

\paragraph{Key Findings.} Our experiments reveal patterns that challenge conventional wisdom: (1) \textbf{truncated reasoning can hurt} as DeepSeek-V3.2 achieves 53\% with no reasoning but only 17\% with truncated CoT at 50\% budget, suggesting incomplete chains actively mislead models; (2) \textbf{code degrades gracefully} as Gemini's comments collapse to 0\% while code maintains 43-47\% at 30-50\% budgets, indicating structured formats have greater truncation tolerance; (3) \textbf{hybrid reasoning underperforms} single modalities, pointing to modality-switching overhead; (4) \textbf{robustness is model-dependent} as Grok maintains 80-90\% at 30\% budget where OpenAI (10-27\%) and DeepSeek (7-47\%) show substantial degradation.

\paragraph{Contributions.} We (1) introduce a controlled framework for isolating reasoning modality effects; (2) conduct the first comprehensive token ablation study across reasoning modalities and frontier models including DeepSeek-V3.2; (3) discover that truncated reasoning can actively harm performance; and (4) establish a robustness hierarchy providing actionable guidance for practitioners deploying reasoning systems under constraints.


\section{Related Work}

\paragraph{Chain-of-Thought Reasoning.}
Chain-of-thoug -ht prompting demonstrated that generating intermediate reasoning steps improves multi-step reasoning in LLMs~\citep{wei2022chain}, with follow-up work showing even simple ``step-by-step'' instructions unlock zero-shot gains~\citep{kojima2022large}. 
CoT has culminated in reasoning-specialized models such as OpenAI's 5.1~\citep{openai2025gpt51} and DeepSeek-V3.2~\citep{deepseekai2025deepseekv32}, which allocate substantial compute to extended reasoning traces during inference. 
These approaches assume explicit reasoning is beneficial. our work stress-tests this assumption under token constraints.

\paragraph{Safety for CoT} Recent work has raised concerns regarding the faithfulness of chain-of-thought reasoning \citep{barez2025chain}, prompting efforts to improve it through structured interventions \citep{swaroop2025frit}. Relatedly, optimizing the calibration and confidence of CoT outputs remains an open challenge \citep{more2025optimizing}. Beyond faithfulness, privacy risks also arise, as models may inadvertently leak private information during reasoning, necessitating methods to sanitize CoT traces \citep{batra2025salt}. Furthermore, recent analysis reveals that reasoning is not confined solely to semantic tokens but is substantially encoded in punctuation patterns, adding a unique dimension to CoT analysis \citep{chauhan2025punctuation}.

\paragraph{Code-Based Reasoning.}
Program-aided language models (PAL) shift computation from the model to an external executor, reducing calculation errors~\citep{gao2022pal}. Program-of-Thoughts (PoT) frames this as disentangling reasoning structure from execution~\citep{chen2022pot}, while tool-augmented methods generalize code generation to broader utility invocation~\citep{schick2023toolformer}. These works demonstrate code's effectiveness but do not systematically compare code versus natural language reasoning under constrained budgets a gap we address.

\paragraph{Token Budgets and Reasoning Efficiency.}
Despite accuracy gains, longer reasoning increases inference cost. Token-budget-aware reasoning proposes dynamically controlling generation length~\citep{han2024tokenbudget}, and recent work suggests more tokens can sometimes hurt when extra generation introduces noise~\citep{wu2024whenmoreisless}. Budget-forcing methods advocate matching reasoning length to task difficulty~\citep{yuan2025abf}. However, prior work does not isolate reasoning \textit{modality} (code vs. language vs. hybrid) as a variable. We provide the first controlled study comparing how different reasoning formats degrade under identical token constraints across multiple frontier models, including reasoning-specialized systems like DeepSeek-V3.2.


\section{Methodology}

\paragraph{Models.} We evaluate four frontier models representing diverse architectures and training approaches: GPT-5.1 (OpenAI), Gemini 3 Flash (Google), DeepSeek-V3.2 (DeepSeek), and Grok 4.1 (xAI). Models were chosen based on high reasoning capability as well as the ability to toggle between reasoning and non-reasoning modes to test non-reasoning capability.

\begin{figure}
    \centering
    \includegraphics[width=\linewidth]{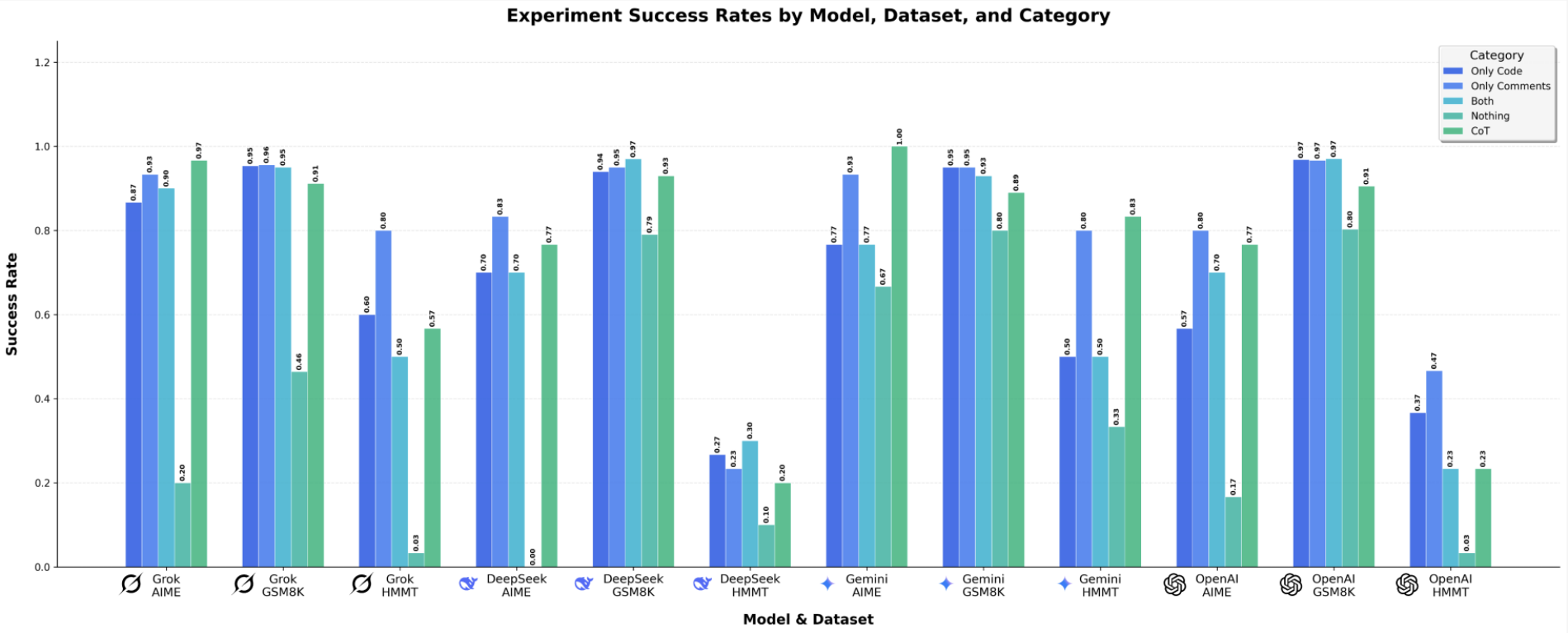}
    \caption{Full performance across all models and datasets. Success rates by model--dataset pair for each reasoning condition (Code-only, Comments-only, Both, Nothing, CoT). Bars show exact-match accuracy under unconstrained generation and tokens.}
    \label{fig:main_results}
\end{figure}

\paragraph{Datasets.} We use three mathematical reasoning benchmarks spanning different difficulty levels: GSM8K~\citep{cobbe2021training} (grade-school mathematics), AIME (American Invitational Mathematics Examination), and HMMT (Harvard-MIT Mathematics Tournament). This range allows us to examine whether reasoning modality effects vary with task difficulty.

\paragraph{Reasoning Conditions.} We constrain each model to reason through five distinct conditions:
\begin{itemize}[leftmargin=*, nosep]
    \item \textbf{Code-only}: Model must express all reasoning as executable code, with the final answer extracted from code output.
    \item \textbf{Comments-only}: Model reasons in natural language comments without executable code.
    \item \textbf{Both}: Model can use both code and comments freely.
    \item \textbf{Nothing}: Model must output the final answer directly without any explicit reasoning.
    \item \textbf{CoT}: Standard chain-of-thought with no modality constraints.
\end{itemize}

Conditions are enforced through carefully designed system prompts that specify the allowed reasoning format and output structure.

\paragraph{Token Ablation.} For each model-dataset-condition combination, we first measure the token count $T_{opt}$ used under unconstrained generation. We then impose token limits at 10\%, 30\%, 50\%, and 70\% of $T_{opt}$, forcing the model to complete its reasoning within the budget. This ablation reveals how gracefully each modality degrades under compression. For DeepSeek-V3.2, which naturally produces lengthy reasoning traces, this ablation is particularly informative.

\paragraph{Evaluation.} We measure success rate as the fraction of problems where the model produces the correct final answer. For code-based conditions, answers are extracted from code execution output. For all conditions, we use exact match against ground truth after normalization.


\section{Results}

\begin{figure}
    \centering
    \includegraphics[width=0.7\linewidth]{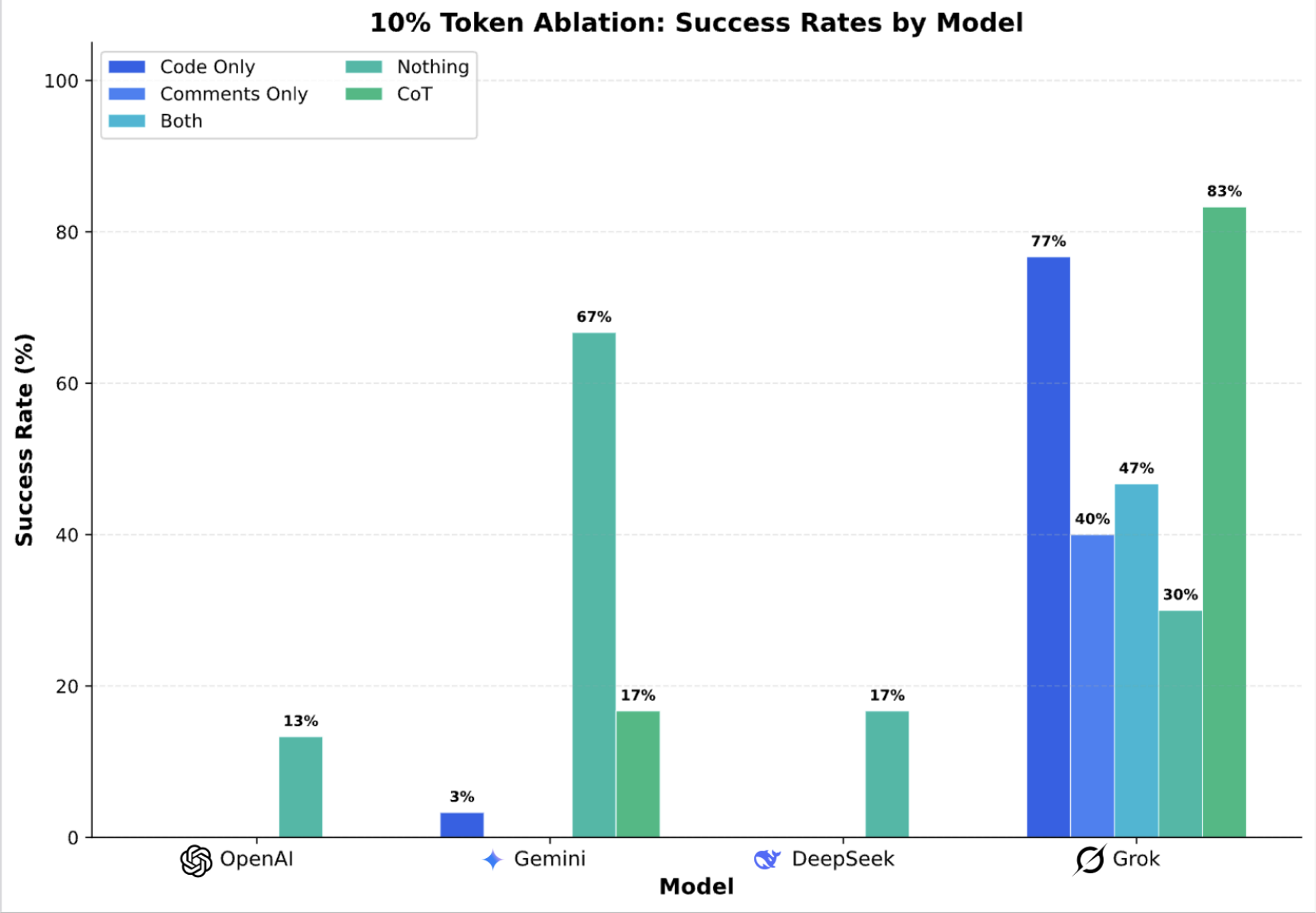}
    \caption{Token-budget ablation for 10\% of the per-setting optimal token count. The x-axis shows the models, and the y-axis shows the success rates given a proportion of token ablation.}
    \label{fig:ablation1}
\end{figure}

\paragraph{Full Token Budget Performance.} Figure~\ref{fig:main_results} presents success rates across all models, datasets, and reasoning conditions without token constraints. Several patterns emerge. First, performance varies substantially across datasets: GSM8K yields highest accuracy (79-100\%) while HMMT proves most challenging (3-83\%). Second, the \textit{nothing} condition shows that explicit reasoning is essential for most models accuracy drops to 3-10\% on HMMT for OpenAI, Grok, and DeepSeek when models cannot reason explicitly, though Gemini maintains 50\%. Third, comments-only reasoning matches or exceeds CoT in most configurations, with Gemini achieving 100\% on GSM8K with comments versus 89\% with CoT.

\paragraph{Token Ablation Reveals Fragility.} Our ablation study (Figure~\ref{fig:ablation1}) exposes how reasoning modalities degrade under token constraints. The most striking finding is that \textbf{truncated reasoning can perform worse than no reasoning}. At 50\% token budget, DeepSeek-V3.2 achieves 53\% accuracy with the \textit{nothing} condition but only 17\% with truncated CoT. Similarly, at 30\% budget, DeepSeek's \textit{nothing} condition (47\%) substantially outperforms truncated CoT (7\%). This pattern is particularly notable given that DeepSeek-V3.2 is explicitly designed for extended reasoning, yet truncating its reasoning traces proves actively harmful. The full ablation study can be found in the appendix (Figure~\ref{fig:ablation}).

\paragraph{Code Exhibits Greater Truncation Tolerance.} Under token constraints, code-based reasoning degrades more gracefully than natural language. At 30\% budget, Gemini's comments-only condition collapses to 0\% while code-only maintains 43\%. At 50\% budget, this gap persists (0\% vs 47\%). We hypothesize that code's syntactic structure preserves semantic coherence even when truncated, a partial loop or conditional still conveys algorithmic intent, whereas truncated natural language may lose critical logical connectives.

\paragraph{Hybrid Reasoning Underperforms.} Contrary to intuition, combining code and comments (\textit{both} condition) never achieves best performance. Across all token budgets, single-modality reasoning outperforms hybrid. At 30\% budget, Grok achieves 90\% with code-only and 90\% with comments-only, but only 80\% with both. This suggests modality-switching incurs overhead as tokens spent transitioning between formats provide less reasoning value than sustained single-modality computation.

\paragraph{Model-Dependent Robustness.} Perhaps our most significant finding is the substantial heterogeneity in reasoning robustness across models. Grok maintains 80-90\% accuracy at 30\% token budget across code, comments, and CoT conditions, while OpenAI (10-27\%) and DeepSeek (7-47\%) show severe degradation at the same budget. At the extreme 10\% budget, Grok still achieves 77\% with code-only and 83\% with CoT, whereas OpenAI and DeepSeek collapse to near-zero on reasoning conditions.


\section{Discussion}

\paragraph{Why Does Truncated Reasoning Hurt?} Our finding that incomplete reasoning chains underperform no reasoning challenges the assumption that partial reasoning provides partial benefit. We hypothesize that truncated chains leave models in inconsistent intermediate states causing variables declared but not resolved, logical premises established but not concluded. When forced to produce an answer from this state, models may hallucinate completions that contradict the partial reasoning, whereas direct answering bypasses this failure mode entirely. This effect is pronounced in DeepSeek-V3.2, possibly because its training reinforces dependence on complete reasoning trajectories.

\paragraph{Why Is Code More Robust?} Code's structural properties may explain its truncation tolerance. Programming constructs carry semantic meaning even when incomplete, a partial \texttt{for} loop still signals iteration, a truncated conditional still indicates branching logic. Natural language lacks this structural redundancy; truncating ``the answer is therefore'' provides no information about what follows. Additionally, code encourages decomposition into discrete steps, each somewhat self-contained, while natural language reasoning often builds long dependency chains.

\paragraph{Why Does Hybrid Underperform?} Mixing modalities appears to incur coordination costs. Transitioning from code to comments requires the model to (1) summarize computational state in natural language, then (2) re-encode this summary for subsequent code. These translations consume tokens without advancing the solution. Single-modality reasoning avoids this overhead, dedicating all tokens to forward progress.

\paragraph{Implications for Reasoning-Specialized Models.} Our results raise questions about the robustness of reasoning-specialized models like DeepSeek-V3.2 and 5.1. If these models are trained to produce extended reasoning traces, they may become \textit{dependent} on completing those traces, making them vulnerable in deployment scenarios with token constraints. Practitioners should consider whether direct answering or code-based reasoning might outperform truncated CoT in latency-sensitive applications.


\section{Conclusion}

We presented a systematic study of reasoning modality robustness under token constraints. Our experiments across four frontier models including the reasoning-specialized DeepSeek-V3.2 reveal that truncated chain-of-thought can actively harm performance, code-based reasoning degrades more gracefully than natural language, hybrid reasoning consistently underperforms, and robustness varies dramatically across models.

\section{Limitations} Our study focuses on mathematical reasoning; generalization to other domains (coding, commonsense, scientific reasoning) remains to be verified. Additionally, our token ablation uses fixed percentages rather than adaptive budgets.

\paragraph{Future Work.} Promising directions include investigating \textit{why} Grok exhibits such robustness (architectural analysis), developing adaptive token allocation strategies that select modality based on problem difficulty, and examining whether reasoning-specialized models like DeepSeek-V3.2 and 5.1 can be fine-tuned for truncation robustness without sacrificing full-budget performance.


\bibliography{main}

\section{Appendix}
\label{app:appendix}
\subsection{Dataset Details}

\begin{figure}[h]
\centering
\begin{minipage}{0.48\textwidth}
    \centering
    \includegraphics[width=\linewidth]{figures/10pct.pdf}
\end{minipage}\hfill
\begin{minipage}{0.48\textwidth}
    \centering
    \includegraphics[width=\linewidth]{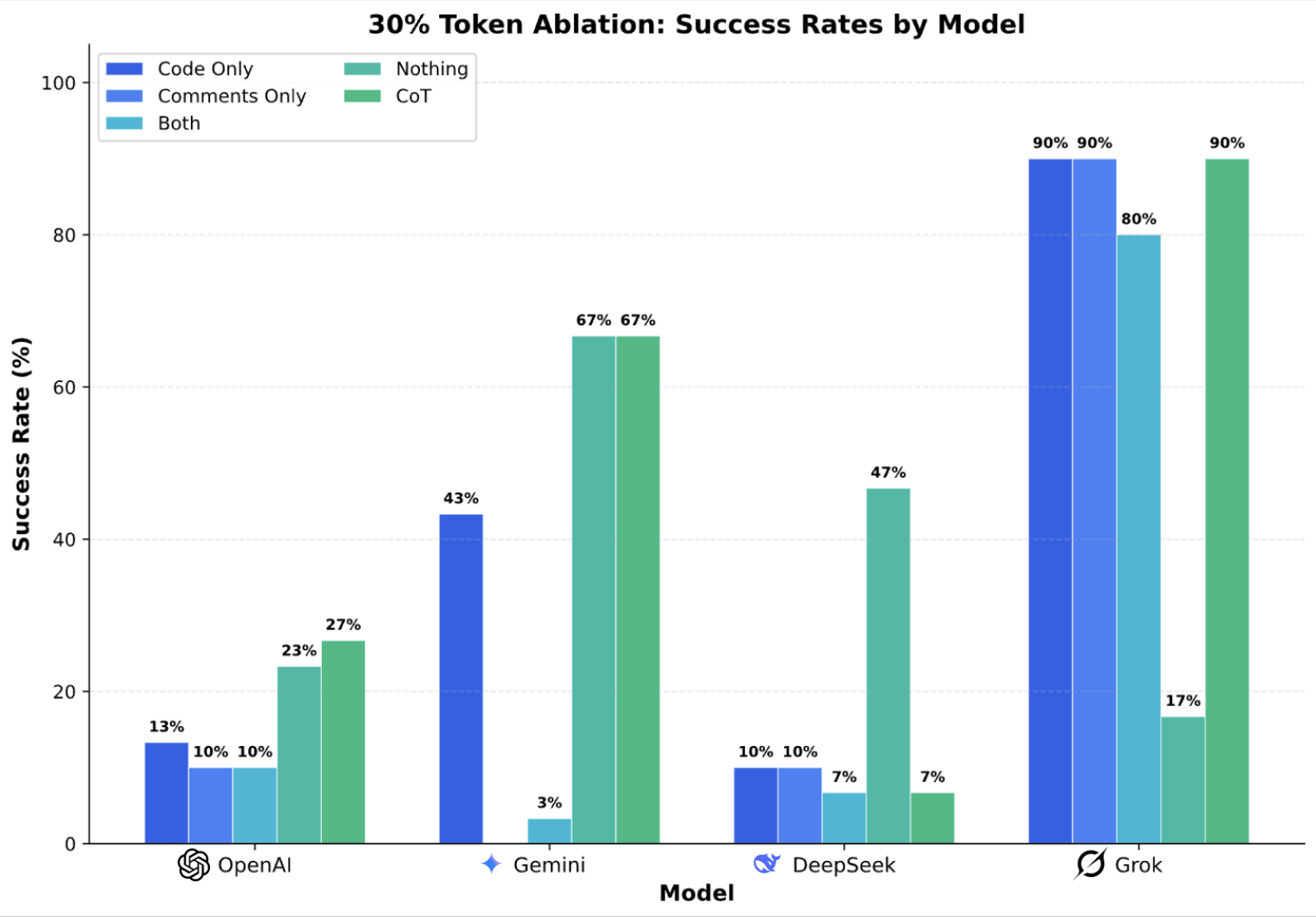}
\end{minipage}
\begin{minipage}{0.48\textwidth}
    \centering
    \includegraphics[width=\linewidth]{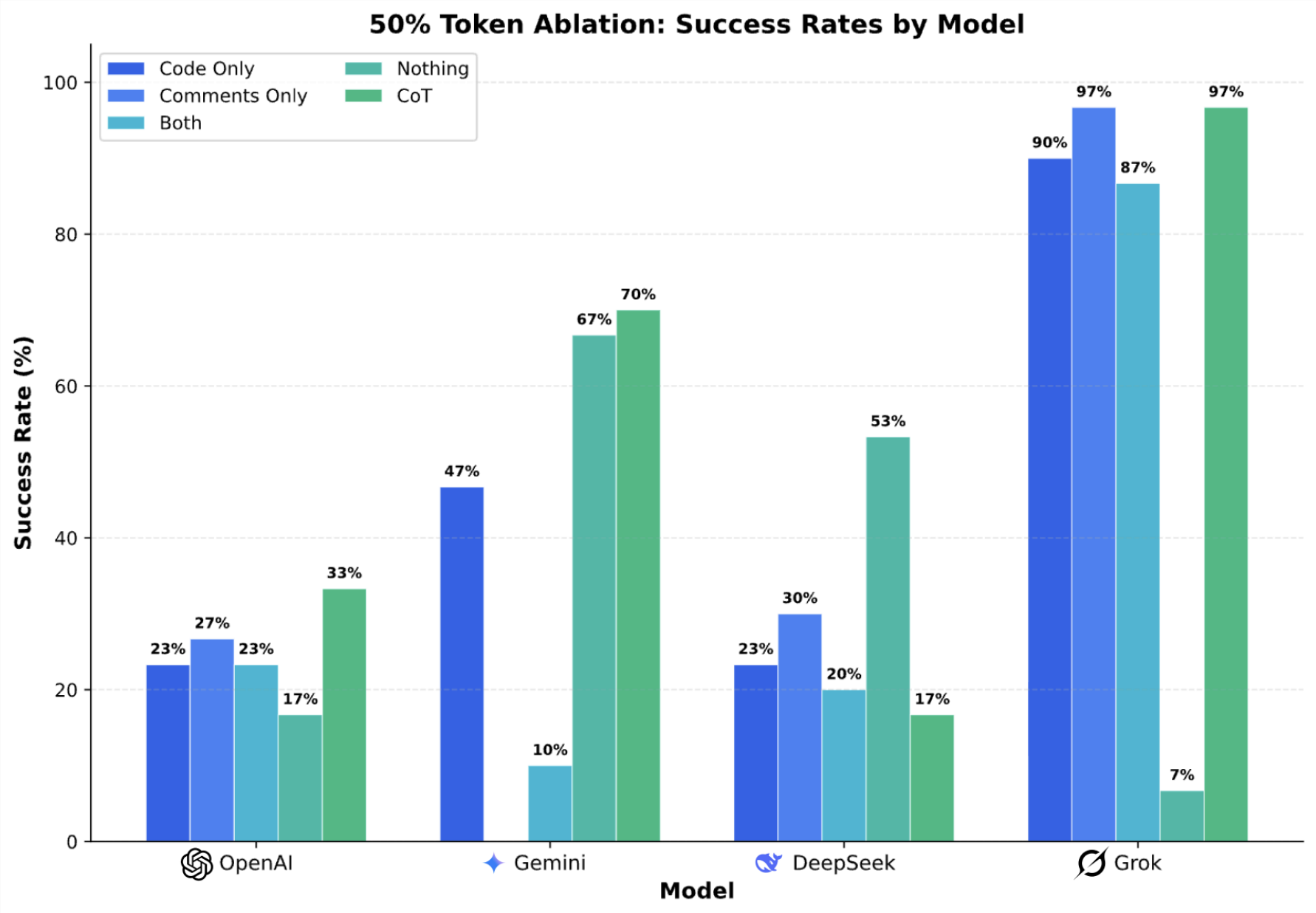}
\end{minipage}
\begin{minipage}{0.48\textwidth}
    \centering
    \includegraphics[width=\linewidth]{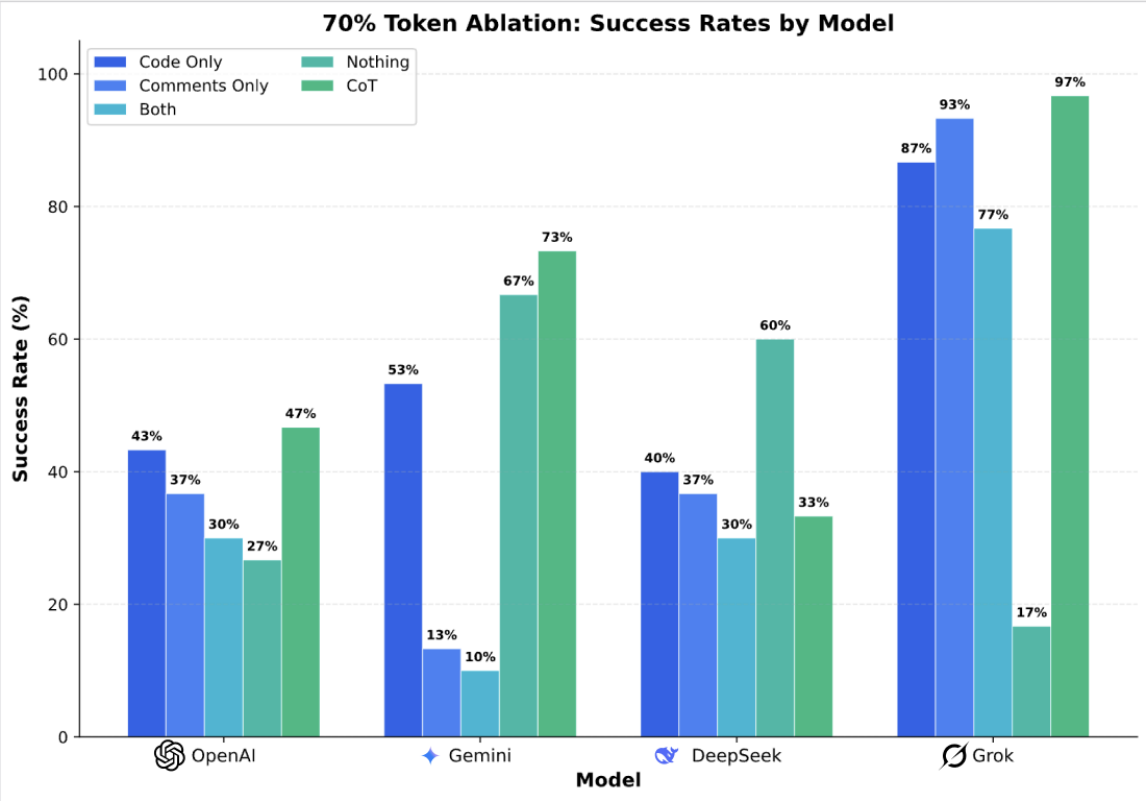}
\end{minipage}
    \caption{Token-budget ablation for 10\%, 30\%, 50\%, and 70\% of the per-setting optimal token count. The x-axis shows the models, and the y-axis shows the success rates given a proportion of token ablation.}

    \label{fig:ablation}
\end{figure}

We evaluated four frontier models (Grok, OpenAI, Gemini, DeepSeek) on AIME problems under token budget constraints of 10\%, 30\%, 50\%, and 70\%, testing code-only, comment-only, combined, no-reasoning, and chain-of-thought conditions.
Results revealed dramatic differences in robustness to truncation. At 10\% tokens, OpenAI and DeepSeek failed entirely (0\% accuracy), while Grok maintained 83\% CoT accuracy. We observed a "truncation paradox" where no reasoning outperformed truncated reasoning for some models—Gemini achieved 67\% with no reasoning versus 17\% with truncated CoT, suggesting incomplete reasoning chains can actively mislead certain architectures. Grok consistently dominated across all conditions, reaching 97\% CoT accuracy at 70\% tokens.

\subsection{Dataset Details}






\end{document}